\documentclass[letterpaper]{article} 
\usepackage{aaai25}  
\usepackage{times}  
\usepackage{helvet}  
\usepackage{courier}  
\usepackage[hyphens]{url}  
\usepackage{graphicx} 
\usepackage{natbib}  
\usepackage{caption} 
\usepackage{algorithm}
\usepackage{algorithmic}

\usepackage{newfloat}
\usepackage{listings}
\DeclareCaptionStyle{ruled}{labelfont=normalfont,labelsep=colon,strut=off} 
\floatstyle{ruled}
\newfloat{listing}{tb}{lst}{}
\floatname{listing}{Listing}
\title{A Weak Supervision Learning Approach Towards an Equitable Mobility Estimation}
\author {
    Theophilus Aidoo,\textsuperscript{\rm 1}
    Till Koebe\textsuperscript{\rm 1},
    Akansh Maurya\textsuperscript{\rm 1},
    Hewan Shrestha\textsuperscript{\rm 1},
    Ingmar Weber\textsuperscript{\rm 1}
}
\affiliations {
    \textsuperscript{\rm 1} Department of Computer Science, University of Saarland\\
    taidoo@cs.uni-saarland.de, tkoebe@cs.uni-saarland.de, akma00001@uni-saarland.de, hesh00001@uni-saarland.de, iweber@cs.uni-saarland.de
}

\usepackage{bibentry}

\begin{document}

\maketitle

\begin{abstract}\footnote{To appear in the proceedings of the ICWSM'25 Workshop on Data for the Wellbeing of Most Vulnerable (DWMV). Please cite accordingly.}
The scarcity and high cost of labeled high-resolution imagery have long challenged remote sensing applications, particularly in low-income regions where high-resolution data are scarce. In this study, we propose a weak supervision framework that estimates parking lot occupancy using 3m resolution satellite imagery. By leveraging coarse temporal labels—based on the assumption that parking lots of major supermarkets and hardware stores in Germany are typically full on Saturdays and empty on Sundays — we train a pairwise comparison model that achieves an AUC of $0.92$ on large parking lots. The proposed approach minimizes the reliance on expensive high-resolution images and holds promise for scalable urban mobility analysis. Moreover, the method can be adapted to assess transit patterns and resource allocation in vulnerable communities, providing a data-driven basis to improve the well-being of those most in need.

\end{abstract}

\section{Introduction}
\label{sec:intro}
Unlike image classification tasks in natural images, where extensive labeled datasets such as ImageNet \cite{deng2009imagenet} and MS COCO \cite{lin2014microsoft} are readily available, remote sensing applications suffer from a lack of large-scale labeled data. This shortage hinders the development of broadly applicable representations for tasks like universal parking lot estimation across the globe. The scarcity of labels is largely due to the prohibitively expensive and labor-intensive nature of manual annotation \cite{christie2018functional, uzkent2019learning, berg2022self, 9751593, rufener2024estimation}. Moreover, a simple survey of image availability reveals that high-resolution captures are far more common for high-income countries than for low-income ones. For example, using UP42’s API console\footnote{https://up42.com}, we obtained 27 images for Berlin, Germany and only 15 for Accra, Ghana, under similar conditions (cloud cover below $20$\% and $30$cm resolution) between 2020-01-01 and 2022-12-31. This observation supports the point made by \citet{engstrom2022poverty} that high-resolution satellites tend to underrepresent low-income countries.

Parking-lot occupancy serves as a scalable, proxy for human presence and movement patterns in urban environments \cite{rufener2024estimation}. Since the vast majority of daily trips involve personal vehicles, fluctuations in parking lot usage directly reflect commuting flows, and commercial activity. By predicting occupancy dynamics of parking lots, we can estimate human mobility trends on a global scale.

Given these challenges in obtaining high-quality labels and high-resolution imagery for low-income countries, alternative methods that require less precise annotations and low-resolution imagery are essential towards the estimation of generalizable parking lot occupancy. Weak supervision learning is one such approach. In this framework, models learn from labels that are inexact (providing only coarse information), incomplete (available for only a subset of the data), or inaccurate (noisy annotations) \cite{zhou2018brief, 9751593, ratner2017snorkel, uzkent2019learning}. Despite their lack of precision, these weak labels can still serve as valuable signals for model training. 

Our work explores weak supervision learning for determining occupied vs empty parking lots using 3m PlanetScope imagery provided by Planet \footnote{\url{https://www.planet.com}} of parking lots in Germany. We aim to contribute to growing research of learning from noisy annotations in domains where ground-truth labels are scarce and expensive to obtain.

\section{Related Works}
\textbf{Weak Supervision.} According to \citet{9751593}, a model is said to be weakly supervised if any of the following conditions hold: (1) there are few labeled instances among many training samples, (2) annotation is provided on a coarse level, or (3) training samples have labels that are only partially correct or noisy. The first condition (1), often referred to as \textit{Incomplete Supervision}, involves learning from a dataset where only a small subset of instances is labeled \cite{he2021online}. In this case, the challenge lies in generalizing from the limited labeled data to the much larger unlabeled portion. This typically involves active learning \cite{zheng2021weakly}, where human experts are used to annotate the unlabeled data points or semi-supervised learning \cite{li2019towards, zheng2021weakly}, where the model learns from both labeled and unlabeled data points.

The second condition (2), \textit{Inexact Supervision} occurs when only coarse labels are available, and the task is to infer finer-grained structures. In remote sensing applications, this often requires interpolation or multi-instance learning techniques to derive detailed insights from coarse data, e.g. identifying individual tree species in a forest from vegetation cover maps \cite{carbonneau2018multiple,wang2022remote,li2023novel}.

The third condition (3), also known as \textit{Inaccurate Supervision}, is a learning paradigm where the annotation of labels may not always reflect the ground-truth \cite{zheng2021weakly}, thus the labels are not accurate or noisy. Noise in the labels can arise from incorrect annotations or flawed assumptions used to generate the labels. This type of supervision poses a unique challenge as models must learn to distinguish between genuine patterns in the data and noise in the labels \cite{8900497}.

\textbf{Parking Lot Occupancy Estimation.} Unlike other satellite imagery providers, PlanetScope provides \emph{near-daily} images, pending cloud-free skies. This high temporal resolution makes this data a potential fit to estimate parking lot occupancy. Using PlanetScope imagery of a parking lot captured on different days, \citet{drouyer2020parking} estimated the occupancy ratio, that is, the proportion of occupied spaces for a large commercial parking lot in the US. The model takes as input a series of images of a parking lot and returns the occupancy ratio.
To achieve this, Drouyer uses three pixel-level features derived from PlanetScope imagery: brightness, gradient, and Pearson correlation with surrounding pixels. These features show strong correlations with ground-truth data captured by a webcam positioned next to the parking lot, with the median image calculated from a series of images when the lot was empty.
In their work on estimating occupancy ratios, \citet{zhao2022vehicle} use a high-resolution image ($30$ cm) for spatial guidance, as cars are easier to identify at that scale. They also use three PlanetScope images: a main image, a second image captured closest in time to the main image, and a third image taken no more than seven weeks apart from the main image.

The pair of images that accompany the main image is used to capture the temporal changes in the images. The use of both high resolution and pair of before/after images yielded a high occupancy correlation to the ground-truth data. The proposed approach achieved a low mean absolute error in vehicle counting, outperforming the baseline models. However, it still requires high-resolution imagery for spatial guidance, which may not be available for retrospective analyses or be expensive to obtain. Another work done in this area is the use of high-resolution SkySat images provided by Planet to label corresponding PlanetScope images as a regression-based task to classify vehicles on the road and further estimate the heading direction of vehicles \cite{van2024vehicle}. 

\section{Our Approach}
\label{sec:formatting}
Existing methods for parking lot occupancy estimation often rely on high-resolution imagery, additional temporal data, which may not always be available or cost-effective or the approach does not generalize well to other parking lots. In contrast, our approach utilizes weak supervision learning, leveraging noisy labels derived from temporal patterns in parking lot usage. This weak supervision arises from the inherent inaccuracy in our labeling assumption—specifically, we assume that parking lots of supermarkets and hardware stores in Germany are typically full on Saturdays and empty on Sundays, as most stores remain closed by national law on Sundays. This policy effect guides our methodology, allowing us to estimate occupancy without requiring extensive labeled data or high-resolution images. While this law and its effects apply widely, there are notable exceptions. For instance, in some communities, these parking lots may host events such as second-hand product sales also known as "flea markets" on Sundays on such parking lots. Similarly, shops may open on special Sundays, known as "verkaufsoffene Sonntag" to compensate for extended holiday closures. Another thing to note is that even though parking lots are expected to be full(er) on Saturdays, due to the popularity of the place, weather condition, and season actual busyness may vary.
In addition to the inaccuracies in our labels, an observer can barely differentiate between subtle features such as individual cars and other structures such as shopping cart stands, garbage containers or other variations in parking lot usage from 3m PlanetScope imagery. Despite these challenges, our goal is to extract general  patterns of presence or absence of cars  from the data by leveraging the weak annotations described above. Once trained we expect to apply our model 'as-is' to new regions without fine-tuning or retraining.

\begin{figure}[!h]
    \centering
    \includegraphics[width=0.45\linewidth]{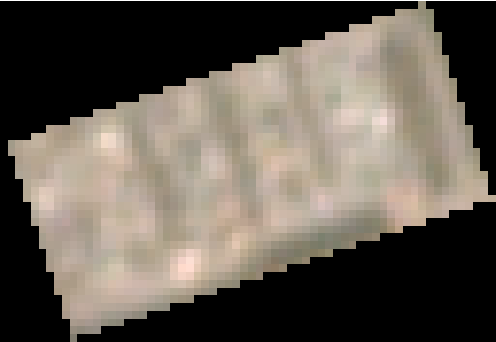}
    \includegraphics[width=0.45\linewidth]{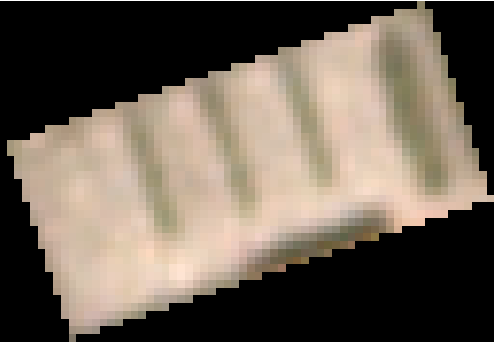}\\
    \includegraphics[width=0.45\linewidth]{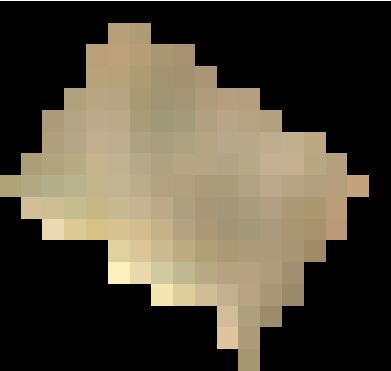}
    \includegraphics[width=0.45\linewidth]{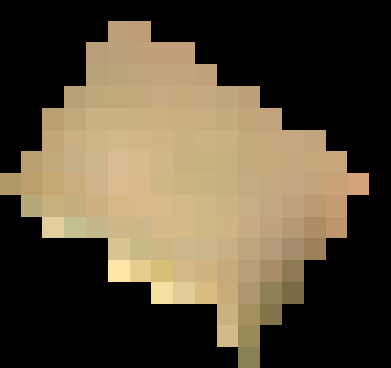}
    \caption{(\textbf{Top row}) A PlanetScope imagery showing the parking situation of a large parking lot observed on a typical Saturday (\textbf{left}), a typical Sunday (\textbf{right}). (\textbf{Bottom row}): A PlanetScope image showing the parking occupancy situation on a small parking lot observed on a Saturday (\textbf{left}), and a Sunday (\textbf{right}).}
    \label{figure:large_small_parkinglots}
\end{figure}

Given a pair of images captured on a Saturday and a Sunday for a large parking lot ($\geq 10,000$ sqm), it is relatively easy for a human observer to determine which image corresponds to an occupied parking lot and which represents an empty one. For instance, the upper row of Figure \ref{figure:large_small_parkinglots} presents a typical example of such an image pair for a parking lot with a total area of $36,866$ sqm. In this case, the image on the left appears more colorful and generally darker in the middle, which can be attributed to the presence of (parked) cars. In contrast, the image on the right exhibits a more homogeneous appearance, indicating an empty parking lot. However, this visual distinction becomes increasingly challenging for smaller parking lots ($\leq 5,000$ sqm), where larger patterns as in large parking lot become less visible. As shown in bottom row of  Figure \ref{figure:large_small_parkinglots}, for a parking lot with an area of $2,971$ sqm, the variation in occupancy between the Saturday and the Sunday image is not as easily discernible. This limitation highlights the need for an automated, scalable approach based on computer vision to determine whether a parking lot is occupied or empty, particularly when visual cues are subtle or ambiguous.

In this work, we set out to overcome expensive labeling efforts usually required for supervised learning tasks. For our study, we used PlanetScope images worth around EUR $3,400$. In comparison, acquiring a similar training dataset (i.e. about 400 parking lots captured on average 48 times across a span of $9$ years, namely $2016$ to $2024$) for estimating parking lot occupancy based on a traditional car detection approach (i.e. high-resolution imagery + car detection model), it would cost between EUR $450,000$ and EUR $500,000$ depending on the vendor and discounts. Apart from the costs, the availability of (historic) high-resolution imagery also limits the number of applications that can be explored.
To address the above challenge, we leverage weak supervision learning to infer occupied and empty parking lots from satellite images captured on Saturdays and Sundays, respectively.
Our contributions in this work include the development of a pipeline for acquiring parking lot polygons from OpenStreetMaps (OSM)\footnote{\url{https://www.openstreetmap.org}} based on Places of Interest. The full processing pipeline is published alongside the paper on GitHub\footnote{\url{https://github.com/Societal-Computing/equitable_mobility_estimation}}.

\subsection{Data}\label{subsec_data}
\begin{figure}[!h]
    \centering
    \includegraphics[width=1\linewidth]{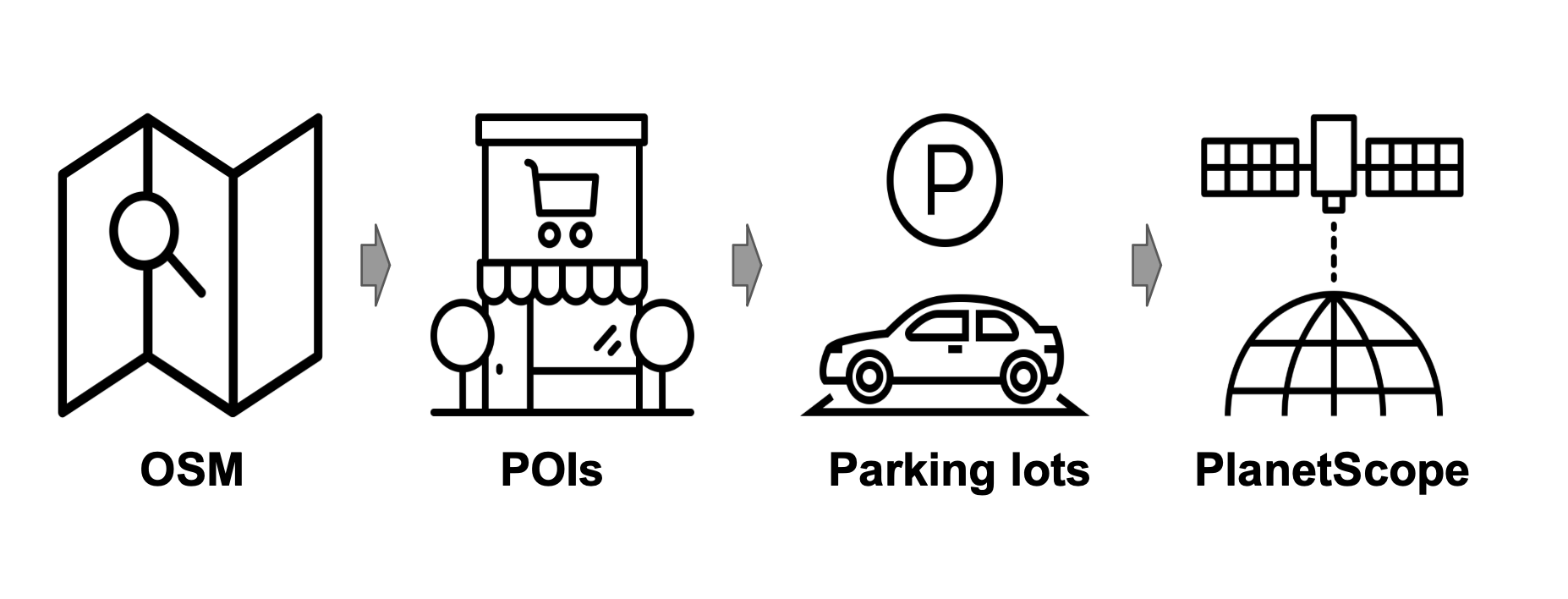}
    \caption{Data acquisition pipeline from POI identification on OSM to acquiring PlanetScope images.}
    \label{fig:data_acq}
\end{figure}

As illustrated in Figure \ref{fig:data_acq}, we first identify Places of Interest (POI), specifically supermarkets and hardware stores (DIY), as these locations typically have dedicated parking lots. For each supermarket or hardware store, we retrieve the nearest parking lot polygon from OSM. In order to ensure that only clearly identifiable parking lots are selected, we also limit our search to rooftop and surface parking lots that are open to customers to reduce the inclusion of irrelevant parking areas, such as private or abandoned parking spaces, which could introduce further noise into the dataset. Additionally, we set a proximity threshold of $10$ m around each POI to exclude distant parking lots that may not be directly associated with the identified POI.
After applying these selection criteria, we identify a total of 745 parking lots across Germany. We obtain both 4-band and 8-band ortho-rectified surface reflectance PScene images using the Planet API for the summer period (April - September), from $2019$ to $2023$ and April - May in $2024$. This season was chosen because cloud cover is generally lower. PlanetScope imagery is captured by a constellation of small satellites at a spatial resolution of $3$m, known as CubeSats, which capture images of the world on a near-daily frequency. However, due to variations in satellite coverage and stitching, some images only partially covered our parking lots, leading to inconsistencies in image dimensions. To address this issue, for a single parking lot, we removed all images that do not fully cover the area of interest.

Furthermore, we utilize the improved Usable Data Mask (UDM2), which accompanies each ortho-rectified analytic PlanetScope image, to filter out images containing cloudy pixels. Additionally, to maintain consistent brightness across images, we remove any image whose histogram distribution differs by more than $0.2$ from that of the median image for its corresponding parking lot. This step helps to mitigate brightness variations introduced by different satellite sensor captures, ensuring uniformity in the dataset.

After these preprocessing steps, we retain $683$ parking lots which include $355$ large parking lots ($\geq 10,000$  sqm) , $142$ medium parking lots (between $5,000$ sqm and $ 10,000$ sqm) and $186$ small parking lots ($\leq 5,000$ sqm).

\subsection{Model}\label{subsec_model}
Our main contribution in this paper is the weak supervision learning approach from the observation made on parking lots of large supermarkets and hardware stores in Germany. We also show a baseline model which we use to evaluate our approach. The processing is illustrated in Figure \ref{fig:model_arch} below.
The data that goes into the model are image pairs, and their corresponding labels. We assign label $1$ for a Saturday-Sunday pair for image 1 and image 2, respectively, and assign label $0$ for a Sunday-Saturday pair for image 1 and image 2, respectively. Our pairwise comparison model consists of two encoders, namely \textbf{Encoder 1} and \textbf{Encoder 2} with weight sharing. Each encoder converts the input image into a $128$ feature vector. Specifically, \textbf{Feature 1} for image 1 and \textbf{Feature 2} for image 2. We then do a simple difference between the the feature vectors and the results serve as an input to a two-layer \textbf{MLP}. The purpose of the MLP is to project the 128 vector from the differences between Feature 1 and Feature 2 to a single output neuron. We then apply a sigmoid function to convert the logit values to probability values.
\begin{figure}[!h]
    \centering
    \includegraphics[width=\linewidth]{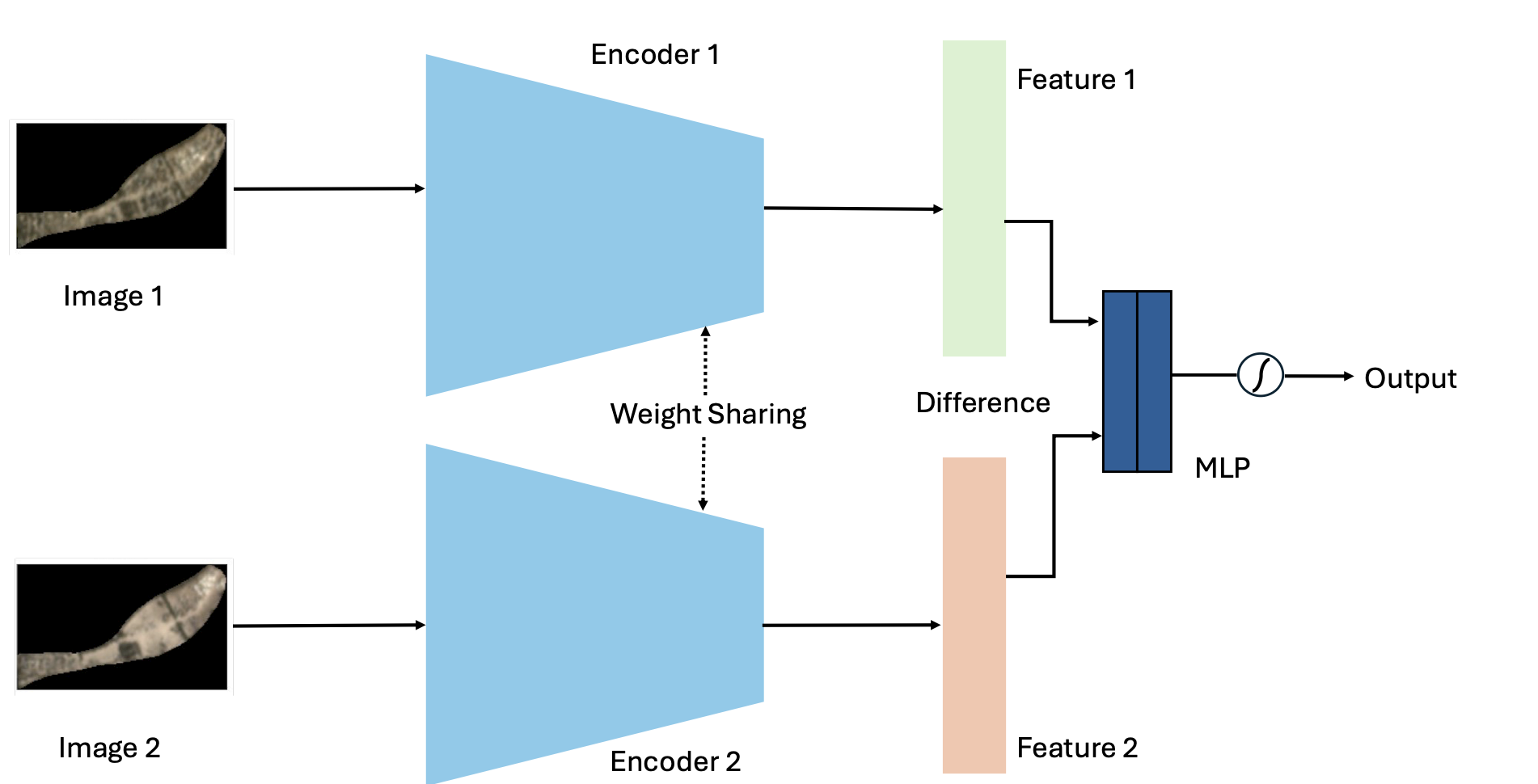}
    \caption{Given a pair of images, \textbf{Encoder 1} and \textbf{Encoder 2} are used to extract features of a lower dimension, yielding \textbf{Feature 1} and \textbf{Feature 2} for image 1 and image 2 respectively. The difference between the feature vectors is then projected to a single output with a Multi-Layer Perceptron (\textbf{MLP}).}
    \label{fig:model_arch}
\end{figure}

We use ResNet50 \cite{he2016deep} as our encoder, based on its high accuracy on remote sensing classification tasks \cite{harini2024resnet} and parking lot estimation tasks \cite{thakur2024deep}, we also used a threshold value of $0.5$ to separate output class $0$ from $1$. We will release a robust comparison with other encoders in future works as our main goal in this paper is to introduce the weak supervision learning approach. Also we consider our pairwise comparison model as a baseline to future approaches.

\section{Experiments}
We perform an $80$-$20$ train-test split at the parking lot-level separately for large, medium, and small parking lots, using the same pairwise comparison approach.
We observe from our results in Table \ref{tab:main_results}, that for large and medium sized parking lots, the model is able to distinguish between occupancy difference between Saturday and Sunday images. However, for small parking lots our approach yields a low AUC score, which we attribute to the difficulty in identifying relatively occupied and empty parking lots of parking lot of small size. We consider a further analysis on small parking lots as a future work. 

\begin{table}[!h]
\centering
\begin{tabular}{l|l}
    Parking lot size & AUC Score \\
    \hline
    large parking lots & \textbf{0.92} \\
    medium parking lots & 0.91 \\
    small parking lots & 0.65 \\
\end{tabular}
\caption{AUC score of pairwise comparison for different parking lot sizes.}
\label{tab:main_results}
\end{table}

In our final experiment, we seek to validate our approach in the context of the on-going conflict between the Sudanese Armed Forces and the Rapid Support Forces that broke out in April 2023. 
As reported by \citet{guo2024monitoring}, there were fewer cars in Sudan during the post\_war period (April 14–21, 2023) compared to the pre\_war period (April 1–7, 2023), as reflected by a decrease in nitrogen dioxide (${NO}_{2}$) concentrations, which are primarily emitted by vehicles and power plants.

\begin{figure}[!h]
    \centering
     \includegraphics[width=\linewidth]{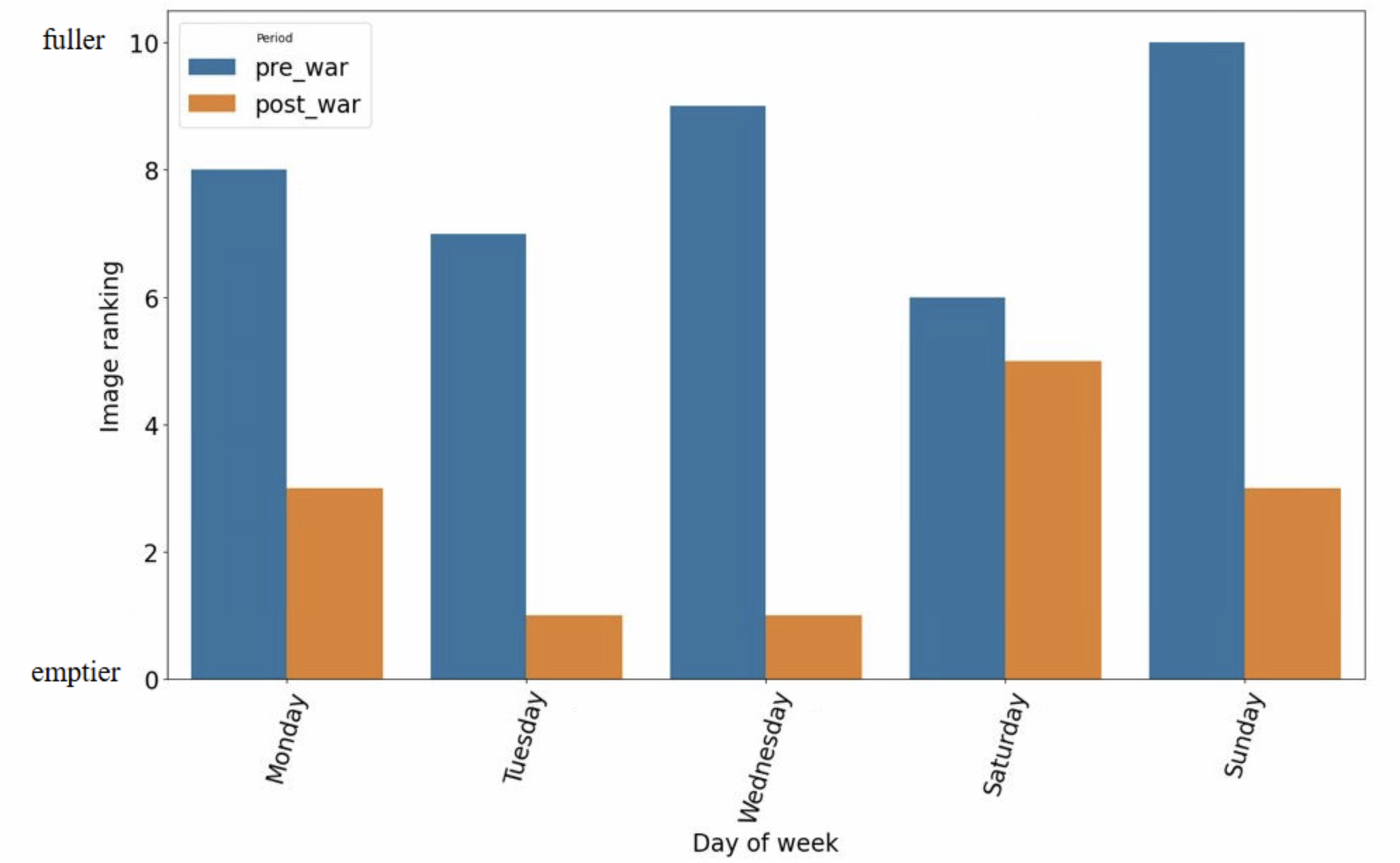}
     \caption{Pairwise comparison of day of week image rankings for Jackson Bus Terminal in Khartoum. The pre\_war period spans 2023-04-01 to 2023-04-07, and the post\_war period spans 2023-04-15 to 2023-04-21. Images captured during the war (orange bars) show consistently lower rankings than their pre-war counterparts (blue bars).}
     \label{fig:sudan_pairs}
\end{figure}

To assess whether our method reflects this observation, we analyzed PlanetScope images of the Jackson Bus Terminal in Khartoum. Specifically, we select days of the week in both the pre\_war and post\_war periods for which images were available, then apply our pairwise comparison approach to all possible image pairs (excluding pairs with the same date). As shown in Figure \ref{fig:sudan_pairs}, images captured during the war period consistently received lower rankings than their pre\_war counterparts, aligning with the observed drop in ${NO}_{2}$ concentrations. 
\section{Conclusion}
In this work, we proposed a weak-supervision learning approach to distinguish relative parking lot occupancies via pairwise comparisons. By exploiting Germany’s Sunday closure law for large supermarkets and hardware stores, we cheaply generated labels (“occupied” vs. “empty” pairs) without manual annotation. Our experiments on large, medium, and small lots show this proxy holds especially well for large and medium facilities. Notably, a model trained solely on large German lots was still able to detect mobility shift of a bus terminal in Sudan during the onset of the war.
Despite the inexactness of our labels, this cost-effective strategy opens the door to many real-world applications, such as detecting population movements around Ukrainian borders, and measuring attendance drop-offs at hospital or market parking lots during epidemics.
In low-income regions where high-resolution data are scarce and public transport hubs or large open-air markets stand in for parking lots, our weakly supervised framework can similarly reveal aggregate human mobility patterns. Future work will extend these case studies and explore real-time deployment for disaster response and humanitarian aid. Overall, our findings demonstrate that even “cheap” weak labels can unlock scalable, global insights into human mobility trends from satellite imagery.
\section{Acknowledgments}
This work is supported by funding from the Alexander von Humboldt Foundation and its founder, the Federal Ministry of Education and Research (Bundesministerium für Bildung und Forschung). We also express our gratitude to the European Space Agency for providing us with UP42 credits to access high-resolution images through the NoR scholarships program.

\bibliography{aaai25}

\end{document}